\pdfoutput=1

\documentclass[11pt]{article}

\usepackage[]{EACL2023}

\usepackage{times}
\usepackage{latexsym}

\usepackage[T1]{fontenc}

\usepackage[utf8]{inputenc}

\usepackage{microtype}
\usepackage{inconsolata}
\usepackage{amsfonts,amsmath}
\usepackage{amssymb}
\usepackage{booktabs}
\usepackage{multirow}
\usepackage{caption}
\usepackage{graphicx}
\usepackage{bm}
\usepackage{subfigure}
\usepackage{paralist, tabularx}
\usepackage{enumerate}
\usepackage{enumitem}
\usepackage{bbding}
\usepackage[nameinlink]{cleveref}

\usepackage{color}
\usepackage{algorithm}
\usepackage{algpseudocode}
\usepackage{mathtools}
\usepackage{arydshln}

\newcommand{\softmax}{\mathop{\mathrm{softmax}}}

\newcommand{\enc}{\mathop{\mathrm{Transformer\_Encoder}}}
\newcommand{\dec}{\mathop{\mathrm{Transformer\_Decoder}}}

\title{Strategize Before Teaching: A Conversational Tutoring System \\ with Pedagogy Self-Distillation}

\author{Lingzhi Wang$^{1,2}$, Mrinmaya Sachan$^3$, Xingshan Zeng$^{4}$, Kam-Fai Wong$^{1,2}$\\
  $^1$The Chinese University of Hong Kong, Hong Kong, China\\
  $^2$MoE Key Laboratory of High Confidence Software Technologies, China\\
  $^3$Department of Computer Science, ETH Zurich \\
  \tt $^{1,2}$\{lzwang,kfwong\}@se.cuhk.edu.hk \\
   \tt $^3$msachan@ethz.ch, \tt $^4$zxshamson@gmail.com
}

\begin{document}
\maketitle
\begin{abstract}
    Conversational tutoring systems (CTSs) aim to help students master educational material with natural language interaction in the form of a dialog. CTSs have become a key pillar in educational data mining research. A key challenge in CTSs is to engage the student in the conversation while exposing them to a diverse set of teaching strategies, akin to a human teacher, thereby, helping them learn in the process. Different from previous work that generates responses given the strategies as input, we propose to jointly predict teaching strategies and generate tutor responses accordingly, which fits a more realistic application scenario. We benchmark several competitive models on three dialog tutoring datasets and propose a unified framework that combines teaching response generation and pedagogical strategy prediction, where a self-distillation mechanism is adopted to guide the teaching strategy learning and facilitate tutor response generation. Our experiments and analyses shed light on how teaching strategies affect dialog tutoring.
\end{abstract}
\section{Introduction}

Decades of research effort \cite{carbonell1970ai,richardson1988foundations,brown2009mobile} has been put into building intelligent tutoring systems (ITSs). An important feature of these systems is the ability to customize the instructional activities and strategies based on the learner’s characteristics and needs \cite{kelecs2009zosmat}. 
Conversational tutoring systems (CTSs) that aim to offer automated tutoring through natural language dialog is a key pillar of ITS research.
Earlier work in CTSs was based on conventional techniques such as Bayesian techniques with rule engines \cite{jeon2010adaptive,weragama2014analysing} and hybrid neural networks \cite{kose2017optimization,stasaski-etal-2020-cima}. While various advanced neural approaches have been applied to open-domain \cite{sordoni2015neural,serban2016building,xing2017topic} and task-oriented dialogue systems \cite{zhao2017generative,lei2018sequicity,peng2020soloist}, conversational tutoring systems have not benefited from the development of these technologies \citep{Macina2023}. 

\begin{figure}[t]
\centering
\subfigure[Two examples of teaching strategy and tutor response]{\label{sfig:introcase_a}
\includegraphics[width=\linewidth]{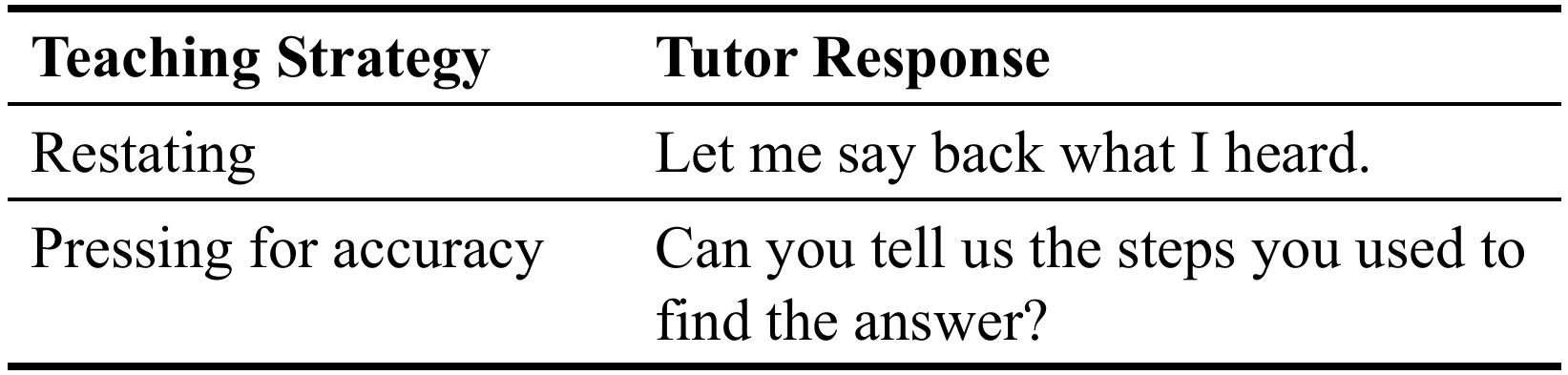}
}
\subfigure[An example of interactions between tutor and student] {\label{sfig:introcase_b}
\includegraphics[width=\linewidth]{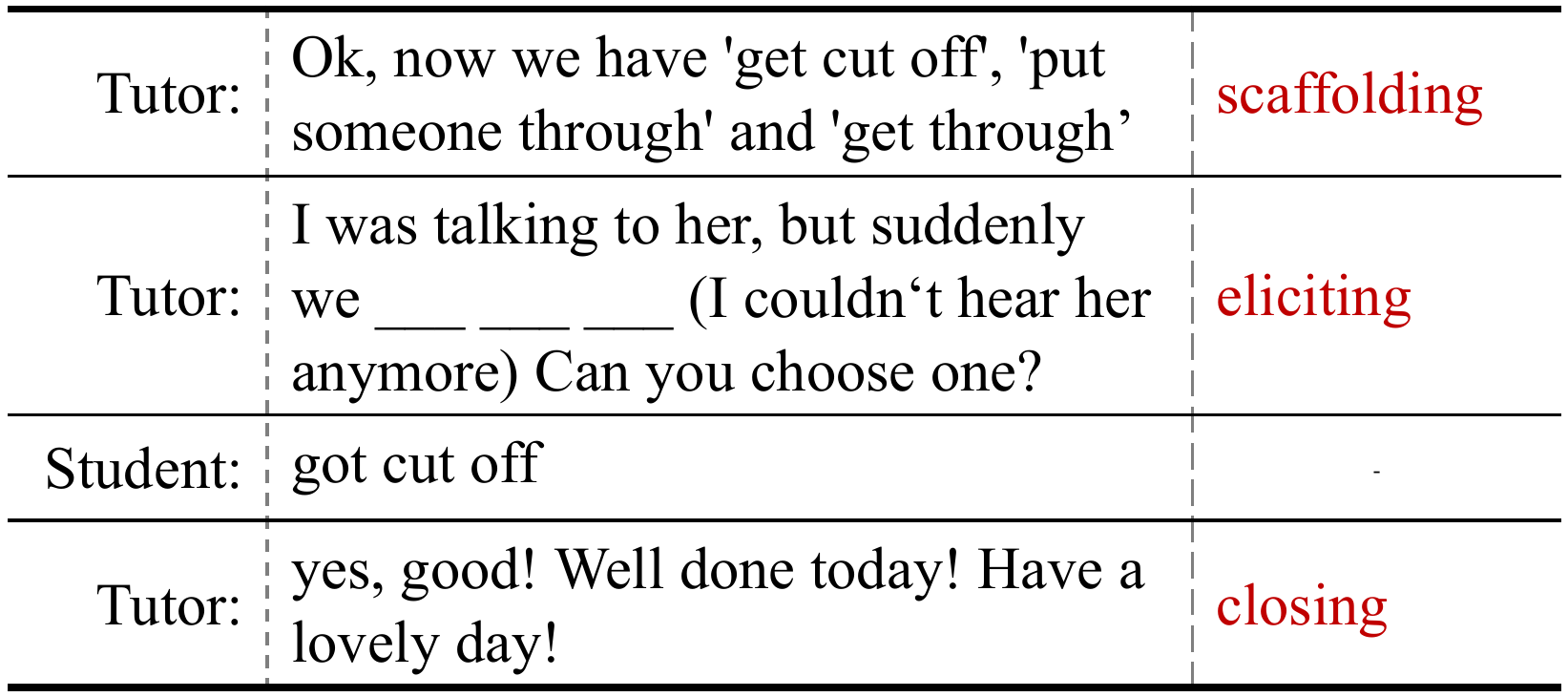}
}
\vskip -0.5em
\caption{\label{fig:introcase}Examples of teaching strategy and interactions between tutor and student. 
Teaching strategies in \Cref{sfig:introcase_b} are in \textcolor{red}{red}.
}
\vskip -1.5em
\end{figure}

Human teachers use a number of nuanced teaching strategies in the classroom during interactions with students; these strategies are tailored to keep the students engaged in the conversation and learn knowledge efficiently. We show some examples of teaching strategies and interactions between the tutor and the student in Fig. \ref{fig:introcase}.
Previous work has attempted to model these teaching strategies in different ways --  e.g., 
\citet{suresh2019automating} contributed a teaching strategy classification model and \citet{stasaski-etal-2020-cima} proposed a response generation model based on given teaching strategies of next response.

In this work, we benchmark several neural dialog models on three conversational tutoring datasets, CIMA~\cite{stasaski-etal-2020-cima}, TSCC~\cite{caines2020teacher} and TalkMoves~\cite{suresh2019automating,suresh2022talkmoves}, and contribute a unified framework based on pretrained language models, where teaching strategy prediction and response generation are jointly trained. As predicting a teaching strategy merely by the historical context is more difficult than when we are also given the target tutor response, we also propose a pedagogy distillation mechanism that allows teaching strategy prediction to learn from the soft labels which are produced by the prediction with target response. 
The soft labels learned from the target response provides the model knowledge about various interrelationships between teaching strategies that hard labels lack. This approach is believed to be able to alleviate the learning difficulty~\cite{hinton2015distilling}, which is particularly important, especially when the data imbalance and scarcity issues are severe -- often the case in conversational tutoring data.

In summary, we are the first to benchmark\footnote{The code can be found in \url{https://github.com/Lingzhi-WANG/TutorSystem}} several competitive models for conversation tutoring system on all three datasets that are currently available. Besides, we propose a unified framework that can predict teaching strategy and generate tutoring responses accordingly, which is enhanced by a self-distillation mechanism. 
Our experiments validate the positive effects of teaching strategy to guide generation and the importance of predicting strategy first and then generate response accordingly.

\section{Related Work}
A classical Intelligent Tutoring System generally has three modules \cite{brown2009mobile,polson2013foundations}: (i) expert module that includes the knowledge that the student wants to learn \cite{carter2014intelligent}). (ii) student module that can adjust the level of student (e.g., primary/middle school, non-native/native speaker), student’s knowledge deficiency, etc. (iii) pedagogical module that focuses on the strategies of teaching.  
In expert module, the knowledge is usually domain specific, such as computer programming \cite{costello2012adaptive}, mathematics \cite{grawemeyer2016affecting,suresh2022talkmoves}, Italian \cite{stasaski-etal-2020-cima}, English \cite{caines2020teacher}. Many technologies have been used in the expert module, such as Bayesian techniques with rule engines \cite{jeon2010adaptive,weragama2014analysing} and hybrid neural networks \cite{kose2017optimization,stasaski-etal-2020-cima}. For pedagogical module, to our best knowledge, there are only three publicly available datasets that provide the pedagogy information. They are CIMA \cite{stasaski-etal-2020-cima}, TSCC \cite{caines2020teacher} and TalkMoves \cite{suresh2022talkmoves} datasets and all of them are based on single pedagogy. 
There has been very little work on neural dialog tutoring. Two exceptions to this are
\citet{suresh2022talkmoves}, who propose a simple BiLSTM-based module to predict the pedagogy of the next sentence that teachers are meant to say, and \citet{stasaski-etal-2020-cima} who use various generative models to generate responses given the pedagogical strategies.
In contrast, in this work, we propose a joint approach for modelling the pedagogy and response generation that outperforms the previous approaches using a novel pedagogy distillation mechanism.

\section{Our Model}
\begin{figure}[t]
\centering
\includegraphics[width=1\linewidth]{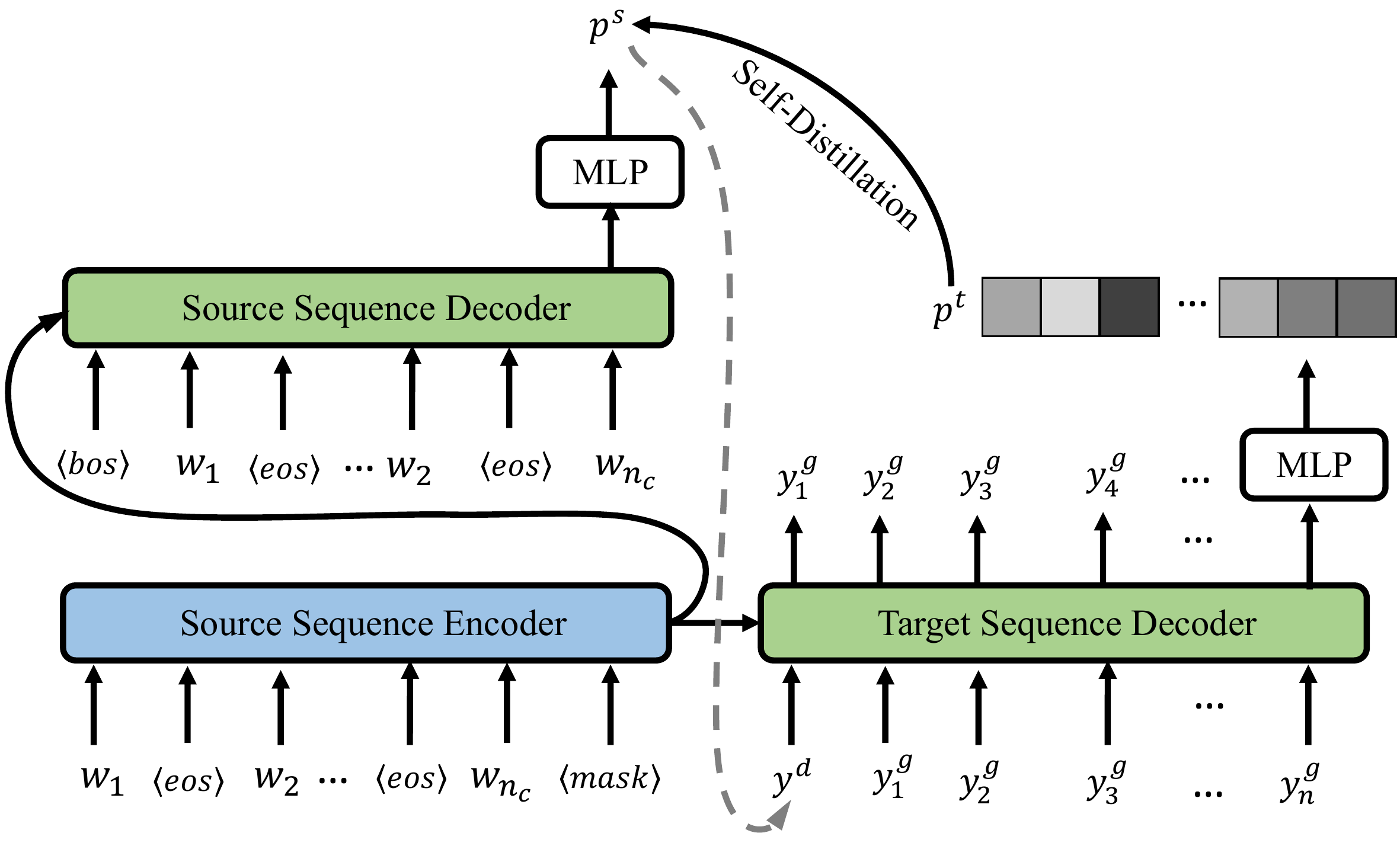}
\vskip -0.5em
\caption{\label{fig:framework} Our overall framework. The self-distillation leverages predictions based on target to improve predictions based on source. The enhanced strategy prediction is further utilized to facilitate the generation. }
\vskip -0.5em
\end{figure}
\subsection{Problem Formulation}
\label{sec:model:form}

Our conversational tutoring system takes conversation context $C$ and teaching strategy list $D$ as input.
$C$ is formalized as a sequence of turns $\{t_1, t_2, ..., t_{n_c}\}$ where $n_c$ represents the number of turns. $t_i$ ($1 \le i \le n_c$) denotes the $i$-th turn of the conversation,
and we use $\bm{w}_i$ to indicate the word tokens contained in it. 
The teaching strategy list $D$ covers all the possible strategies and 
contain $n_d$ teaching strategies.
Our model will first output one or several strategy labels, each $y^d \in \{1,2,...,n_d\}$, to indicate what teaching strategy to use. Then the generation module generates a target response $y^t = (y^t_1, \ldots, y^t_{n_t})$ based on the predicted strategy.

\subsection{Conversational Tutoring System (CTS)}
\label{sec:model:cts}
\paragraph{PLM-based Generation Module.}
The generation module follows a Transformer~\cite{DBLP:conf/nips/VaswaniSPUJGKP17} sequence-to-sequence framework. As the currently available tutoring datasets are quite small (containing about 3k conversations), we choose to finetune pretrained language models (PLM) to alleviate data scarcity and enhance context modeling. We finetune BART \cite{lewis-etal-2020-bart} and multilingual BART(mBART) \cite{liu2020multilingual} models for our generation module.
During finetuning, we concatenate the utterances $t_i$ ($1 \le i \le n_c$) in context $C$ with appended $\langle \text{eos} \rangle$ tokens in their chronological order as input, and maximize the probability of the ground-truth target sequence. The whole process is summarized as follows:
\vskip -1.5em
\begin{align}
\bm{H}^{c} &= \enc(\bm{w}^c)  \\
\label{eq:exp:target_decoder}
y^t_k &= \dec(y^t_{<k},\bm{H}^{c}) \\
\mathcal{L}^{gen}_{target} &= \sum\nolimits_{k=1}^{n_t} - \log (p(y^t_k|y^t_{<k},\bm{H}^{c})) \label{eq:gen_basic}
\end{align}

\noindent where $\bm{w}^c=[\bm{w}_1;\langle \text{eos}\rangle;\bm{w}_2;..;\bm{w}_{n_c};\langle \text{mask}\rangle]$, and $y^t_{<k}$ represents the target tokens before $y^t_k$. 
We add $\langle \text{mask}\rangle$ at the end of context, to simulate the operation in pre-training~\cite{schick2021few}.

Besides, to summarize the representation of the conversation context, we employ an additional source sequence decoder as follows:
\begin{align}
\label{eq:exp:source_decoder}
y^s_k &= \dec(y^s_{<k},\bm{H}^{c})
\end{align}
\begin{align}
\mathcal{L}^{gen}_{source} &= \sum\nolimits_{k=1}^{n_s} - \log (p(y^s_k|y^s_{<k},\bm{H}^{c})) 
\end{align}

\noindent where $y^s_{<k}$ represents the source tokens before $y^s_k$.

\paragraph{Teaching Strategy Prediction Module.}
We use the representation of the $\langle \text{eos}\rangle$ token (i.e. the final token) produced by the decoder as the representation for teaching strategy prediction, denoted as $\bm{h}^{\langle \text{eos}\rangle}$. This is fed into a two-layer MLP for prediction:
\begin{equation}
\bm{r}^d = \bm{W}_2 \times \alpha(\bm{W}_1\bm{h}^{\langle\text{eos}\rangle}  + \bm{b}_1) + \bm{b}_2
\end{equation}
\noindent where $\bm{W}_1$, $\bm{W}_2$, $\bm{b}_1$ and $\bm{b}_2$ are learnable parameters, and $\alpha$ is a non-linear activation function. The output representation $\bm{r}^d$ will be an $n_d$-dimension vector and the probability for each teaching strategy in list $D$ is computed based on $\bm{r}^d$:

\begin{equation}
\label{eq:pred}
p(y^d=j) = \softmax (\bm{r}^d)_j
\end{equation}
where $y^d$ denotes the predicted strategy and $j \in \{1,2,...,n_d\}$.

We denote $\bm{h}^{\langle \text{eos}\rangle}$ produced by source and target generation as $\bm{h}_s^{\langle \text{eos}\rangle}$ and $\bm{h}_t^{\langle \text{eos}\rangle}$, respectively. 
With $\bm{h}_s^{\langle \text{eos}\rangle}$, it means that we predict the teaching strategy without knowing the corresponding content; while with $\bm{h}_t^{\langle \text{eos}\rangle}$, we summarize the teaching strategy based on the target content. Obviously, predicting with $\bm{h}_s^{\langle \text{eos}\rangle}$ is what we need, but this is quite challenging. Thus we design a self-distillation mechanism which uses prediction based on $\bm{h}_t^{\langle \text{eos}\rangle}$ for enhancing the generation model.

\paragraph{Teaching Strategy Enhancement with Distillation.}
We denote the predicted probability for each strategy (derived with Eq.~\ref{eq:pred}) using $\bm{h}_s^{\langle \text{eos}\rangle}$ and $\bm{h}_t^{\langle \text{eos}\rangle}$ as $p_s(\cdot)$ and $p_t(\cdot)$, respectively.
Our self-distillation is defined as guidance from $p_t(\cdot)$ to $p_s(\cdot)$:
\begin{equation}
\label{eq:loss-sd}
\mathcal{L}^{sd} = -  \sum\nolimits_{j=1}^{n_d} p_s(y^d=j) \log p_t(y^d=j)
\end{equation}

\noindent where we define $p_t(\cdot)$ as teacher distribution and $p_s(\cdot)$ as student distribution, and Eq.~\ref{eq:loss-sd} makes the student distribution similar to the teacher distribution.
In this way, our teaching strategy prediction model can also learn from the soft labels produced by the target sequence.

\paragraph{Multiple Teaching Strategies Guided Generation.}
To guide the response generation with teaching strategy, we regard the teaching strategies as prompt tokens and display them at the beginning of generation. In this way, the target tokens will be generated autoregressively according to the giving teaching strategy. Specifically, during training, we use the ground-truth strategy (denoted as $d^c$, and it will be masked in distillation to avoid information leakage) for teacher forcing (i.e. $y^t_0 = d^c$ in Eq.~\ref{eq:gen_basic}); during inference, we use the predicted strategies produced by the prediction module as prompt tokens.

To enable multiple teaching strategies guidance, we define a threshold $\theta$, where all the strategies satisfying $p_s(y^d=j) \ge \theta$ ($1 \le j \le n_d$) will be used to guide the response generation. To that end, we weightedly sum over the embeddings of those strategies as prompt based on their predicted probabilities produced by Eq.~\ref{eq:pred} and then use it to guide the generation.

\subsection{Learning Objectives}
\label{sec:model:objective}
The learning objective for teaching strategy prediction is defined as follows:
\begin{equation}\small
\label{eq:loss_pred}
\mathcal{L}^{pred} = - (\log p_s(y^d=d^c) + \log p_t(y^d=d^c)) + \lambda \cdot \mathcal{L}^{sd}
\end{equation}
where $d^{c}$ is the ground-truth strategy for context $C$ and $\lambda$ is a hyper-parameter to control the weights of self-distillation loss. Our model is jointly trained on both generation and prediction, with the overall objective summarized as:
\begin{equation}
\label{eq:loss_total}
\begin{split}
\mathcal{L} &= \mathcal{L}^{gen} + \gamma \cdot \mathcal{L}^{pred} \\
 &= \mathcal{L}^{gen}_{target} + \delta \cdot \mathcal{L}^{gen}_{source} + \gamma \cdot \mathcal{L}^{pred}
\end{split}
\end{equation}
where $\delta$ and $\gamma$ are tradeoff hyper-parameters.

\begin{table*}[t]
\setlength{\tabcolsep}{2.5mm}
\newcommand{\tabincell}[2]{\begin{tabular}{@{}#1@{}}#2\end{tabular}}
\begin{center}
\resizebox{0.65\linewidth}{!}{
\begin{tabular}{lcccccc}
\toprule
\multirow{2}{*}{\textbf{Models}} & \multicolumn{2}{c}{ \tabincell{c}{\textbf{CIMA}} } & \multicolumn{2}{c}{ \tabincell{c}{\textbf{TSCC}}}
& \multicolumn{2}{c}{ \tabincell{c}{\textbf{TalkMoves}}}
\\
\cmidrule(lr){2-3}\cmidrule(lr){4-5}\cmidrule(lr){6-7}
& Acc & F1 & Acc & F1  & Acc & F1 \\
\midrule
BART &64.3&31.5 &59.1&11.6 &55.2 &31.1 \\
BART$^\dagger$ & \textbf{82.3} & \textbf{57.1} &\textbf{64.4}& \textbf{18.9} & \textbf{75.9} & \textbf{50.5} \\
\midrule
Frequency &62.7&15.4 &58.4&4.1 &52.5 &11.5 \\
BiLSTM &57.3 &30.1 &56.5&11.2 &50.1 &25.6 \\
Transformer &63.3 &33.9 &57.2&16.2 &53.6 &30.7 \\
\midrule
Our Model(BART)&69.7&39.2 & \underline{60.6}&\underline{17.4} &57.8 &35.5\\
Our Model(mBART)& \underline{70.4}& \underline{39.8} &60.4&17.0 &\underline{59.6} & \underline{37.6}\\
\bottomrule
\end{tabular}
}
\end{center}
\vskip -1em
\caption{\label{tab:ts_prediction} Teaching strategy prediction results (in \%). $\dagger$ indicates the prediction is based on the target tutor response. The best and second-best results in each
column are in \textbf{bold} and \underline{underlined} respectively.
}
\vskip -0.5em
\end{table*}

\section{Experimental Setup}
\paragraph{Datasets.}
We use three datasets to do the experiments. They are CIMA \cite{stasaski-etal-2020-cima}, TSCC \cite{caines2020teacher} and TalkMoves \cite{suresh2019automating,suresh2022talkmoves}. CIMA contains one-to-one conversations that focus on teaching students to translate a phrase from English to Italian. TSCC focuses on teaching English for eight non-native English-speaking students. TalkMoves is constructed by transcripts of math classrooms. 

\paragraph{Parameter Setting.} Our implementation is based on Fairseq~\cite{ott2019fairseq}. We split the data into 8:1:1 for training, validation and test. All the hyper-parameters are chosen by grid-search based on the validation performance. 

We use BART-Base\footnote{\url{https://github.com/facebookresearch/fairseq/tree/main/examples/bart}} and mBART-Large\footnote{\url{https://github.com/facebookresearch/fairseq/tree/main/examples/mbart}} models to initialize our model, respectively. 
BART-Base model has $6$ layers of encoder and decoder with $768$ hidden dimension, while mBART-Large has $12$ layers of encoder and decoder with $1024$ hidden dimension. The parameter sizes for the two models initialized with BART and mBART are 199M and 816M, respectively.

We use one NVIDIA RTX 3090 GPU to train our model.
During training, we set the max tokens of each batch to 1024 (for BART, or 512 for mBART) with an update frequency of 4. We adopt Adam optimizer~\cite{kingma:adam} with learning rate selected in $\{$1e-4, 5e-5, 2e-5, 1e-5$\}$
and warm-up updates selected in $\{$200, 500, 1000$\}$ followed by a polynomial decay scheduler.
Dropout strategy~\cite{Srivastava:2014:DSW:2627435.2670313} with dropout rate selected in $\{$0.2, 0.4$\}$ and $L_2$ regularization with 0.01 effect value, as well as early stoping based on validation performance, are used to alleviate overfitting.
We set the tradeoff values among the losses as $\lambda=1.0$, $\gamma=1.0$ and $\delta=0.2$.
During inference, predicting threshold $\theta=0.3$ and beam size is set to 5.

\section{Experimental Results}
\subsection{Teaching Strategy Prediction Results} 
\begin{table*}[t]
\setlength{\tabcolsep}{2.5mm}
\newcommand{\tabincell}[2]{\begin{tabular}{@{}#1@{}}#2\end{tabular}}
\begin{center}
\resizebox{0.8\linewidth}{!}{
\begin{tabular}{p{0.03\textwidth}lcccccc}
\hline
& \multirow{2}{*}{\textbf{Models}} & \multicolumn{2}{c}{ \tabincell{c}{\textbf{CIMA}} } & \multicolumn{2}{c}{ \tabincell{c}{\textbf{TSCC}}}
& \multicolumn{2}{c}{ \tabincell{c}{\textbf{TalkMoves}}}
\\
\cmidrule(lr){3-4}\cmidrule(lr){5-6}\cmidrule(lr){7-8}
& & BLEU & BERT& BLEU & BERT& BLEU & BERT\\
\toprule
\multirow{4}{*}{\parbox[t]{2mm}{\rotatebox[origin=c]{90}{W/O TS}} $\begin{dcases*} \\ \\ \\ \end{dcases*}$}
& BiLSTM &9.08&\textbf{72.6} &1.04&69.0 &0.43&73.2\\
& Transformer &10.1&72.2 &1.53&70.4 &0.74&74.9\\
& BART &6.77& 71.9 &1.27&\textbf{71.2}  &0.85&78.0 \\
& mBART & \textbf{10.6}&70.9  &\textbf{1.96}&68.6  &\textbf{2.95}&\textbf{78.1} \\
\midrule
\multirow{6}{*}{\parbox[t]{2mm}{\rotatebox[origin=c]{90}{With Golden TS}} $\begin{dcases*} \\ \\ \\ \\ \\ \end{dcases*}$}
& BiLSTM &8.61&71.8  &1.32&69.1  &1.42&75.8 \\
& Transformer &11.2&72.8  &1.99&69.9   &2.35&77.4 \\
& BART &9.17&70.8  &1.47&68.6  &2.93&78.0 \\
& mBART &11.1&72.3  &1.57&69.5  &3.38&75.7 \\
\cdashline{2-8}
& \textbf{Our Model(BART)}&10.8&71.4  &2.02&70.6  &3.18&78.0 \\
& \textbf{Our Model(mBART)}&\textbf{12.1}&\textbf{73.8}  &\textbf{2.93}&\textbf{72.6}  &\textbf{5.47}&\textbf{79.7} \\
\midrule
\multirow{6}{*}{\parbox[t]{2mm}{\rotatebox[origin=c]{90}{Need TS Predict}} $\begin{dcases*} \\ \\ \\ \\ \\ \end{dcases*}$}
& BiLSTM &7.65&69.8  &0.68&68.2  &0.48&74.7  \\
& Transformer &8.04&68.6  &0.79&69.3   &2.05&76.8 \\
& BART &7.64&69.5  &1.13&69.4   &1.49&73.8 \\
& mBART &7.77&70.2  &1.57&69.7  &2.44&77.1 \\
\cdashline{2-8}
& \textbf{Our Model(BART)}&8.67&70.8  &2.83&70.0  &2.22&77.5 \\
& \textbf{Our Model(mBART)}&\textbf{11.9}&\textbf{73.0}  &\textbf{2.98}&\textbf{71.9}  &\textbf{4.51}&\textbf{78.6} \\
\bottomrule
\end{tabular}
}
\end{center}
\vskip -1em
\caption{\label{tab:main_gen} 
Generation results (in \%). The best results in each setting are in \textbf{bold}. Our full model achieves significantly better performance than the baselines with the same architecture in the same settings (paired t-test $p<0.05$).
}
\end{table*}
We report the accuracy and Macro F1 scores for teaching strategy prediction task in Table \ref{tab:ts_prediction}. We can find that prediction based on the target tutor response performs much better than merely on source context (comparing BART$^\dagger$ and BART), which indicates that prediction with target content is much easier and also validates our motivation of the self-distillation mechanism. 
With the help of our proposed distillation mechanism, our models with pretrained BART or mBART achieve the best performance in the prediction based on source context. 

\subsection{Tutor Response Generation Results}
We then report case-sensitive detokenized sacreBLEU~\cite{post-2018-call} and BERTScore~\cite{zhang2019bertscore} for tutor response generation in Table \ref{tab:main_gen}.

\paragraph{Three Evaluation Settings.} We show results in three settings in Table \ref{tab:main_gen}. ``W/O TS'' means we don't include teaching strategy (TS) labels in training and testing. ``With Golden TS'' means providing ground truth TS labels for training and testing. ``Need TS Prediction'' means models have to predict TS labels in testing and generate the follow-up tutor responses based on the predicted TS labels.

\paragraph{Analysis on Generation Results.} From Table \ref{tab:main_gen}, we can draw the following main observations.

$\bullet$~\textit{Teaching strategy shows positive effects in generation.}
By comparing the results in ``W/O TS'' and ``With Golden TS'' settings, we observe that guidance from golden teaching strategies improves the generation performance in general, which validates the effects of teaching strategy in guiding generation. Besides, our models further improve their corresponding baselines (e.g. Our Model(BART) v.s. BART), which should result from the joint learning of generation and strategy prediction.

$\bullet$~\textit{Successful guidance requires accurate teaching strategies.}
By comparing results in ``With Golden TS'' and ``Need TS Predict'', we can find that most of the models perform worse when they need to predict strategies first, especially for the baselines with poor strategy prediction performance (refer to results of BiLSTM and Transformer in Table \ref{tab:ts_prediction}). This shows that guidance from inappropriate strategies might even hurt performance, which raises the need for accurate prediction in real-world applications and our proposed method can alleviate the gap significantly.

\subsection{Effects of Teaching Strategy}
\begin{figure}[t]
\centering
\includegraphics[width=1\linewidth]{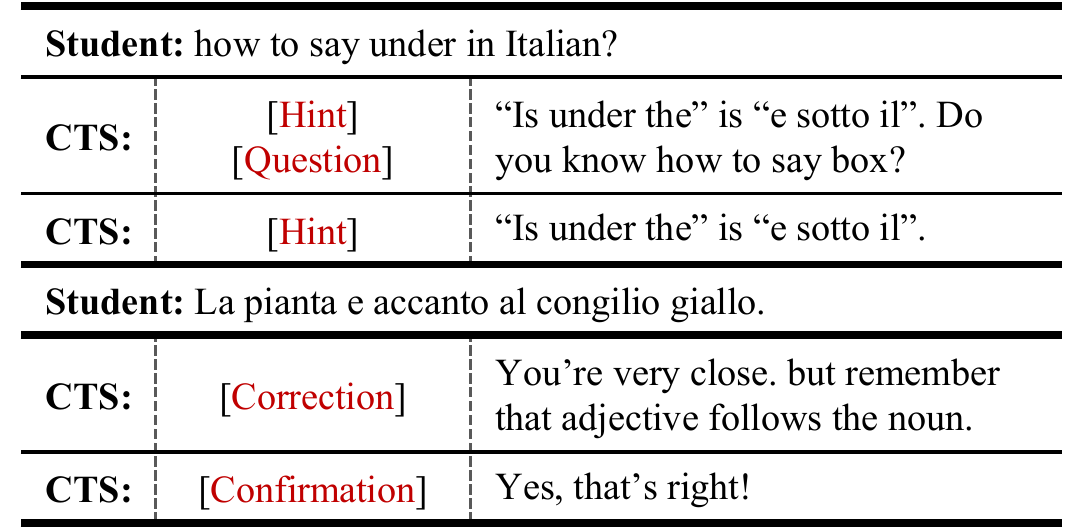}
\caption{\label{fig:da-effects} Our CTS generates different responses when giving different teaching strategies (in \textcolor{red}{red}).}
\end{figure}
We explore how teaching strategy affects the generation in Fig. \ref{fig:da-effects}. We feed our conversational tutoring system (CTS) with different teaching strategies and find that CTS generates totally different responses regarding the same context input. This also validates that teaching strategy is important for a CTS and strategizing before teaching is also essential.
\section{Conclusion}
In this work, we benchmarked neural models on various conversational tutoring datasets and proposed a self-distillation based model that jointly trains a teaching strategy prediction model and a response generation model. Experiments on three conversational tutoring datasets show that our model outperforms various standard baselines by a significant margin. Finally, we ended with an interesting case study to demonstrate the importance of strategizing before teaching.

\section*{Limitations}
There are only three publicly available datasets (CIMA, TSCC and TalkMoves) for conversational tutoring task and they are quite small (less than 10K instances). There are significant data imbalance problems in these datasets -- some teaching strategies occur much more frequently than others. These small and imbalanced datasets bring a lot of challenges to this task, but we did not discuss these issues in our paper due to the space limit. Besides, there are no standard teaching strategy annotation schemes, which prevents us from combining these three datasets together for more interesting experimental analyses. Another limitation of our work is that we only evaluate our approaches on automatic generation metrics. In the future, it would be interesting to also evaluate the model on learning related evaluations. 


\bibliography{anthology,custom}
\bibliographystyle{acl_natbib}

\end{document}